\documentclass[conference]{IEEEtran}
\IEEEoverridecommandlockouts

\usepackage{cite}
\usepackage{amsmath,amssymb,amsfonts}
\usepackage{algorithmic}
\usepackage{graphicx}
\usepackage{textcomp}
\usepackage{xcolor}
\usepackage{url}
\def\BibTeX{{\rm B\kern-.05em{\sc i\kern-.025em b}\kern-.08em
    T\kern-.1667em\lower.7ex\hbox{E}\kern-.125emX}}

\begin{document}
\title{Performance Enhancement Leveraging Mask-RCNN on Bengali Document Layout Analysis\\

}

\author{\IEEEauthorblockN{Shrestha Datta}
\IEEEauthorblockA{\textit{Computer Science and Engineering} \\
\textit{Shahjalal University of Science and Technology}\\
Sylhet, Bangladesh \\
shresthadatta910@gmail.com}
\and
\IEEEauthorblockN{Md Adith Mollah}
\IEEEauthorblockA{\textit{Computer Science and Engineering} \\
\textit{Shahjalal University of Science and Technology}\\
Sylhet, Bangladesh \\
adibhasan35@gmail.com}
\and
\hspace{2cm}\IEEEauthorblockN{Raisa Fairooz}
\hspace{2cm}\IEEEauthorblockA{\hspace{2cm}\textit{Computer Science and Engineering} \\
\hspace{2cm}\textit{Shahjalal University of Science and Technology}\\
\hspace{2cm}Sylhet, Bangladesh \\
\hspace{2cm}raisafairoozshafa@gmail.com}
\and
\hspace{2cm}\IEEEauthorblockN{Tariful Islam Fahim}
\hspace{2cm}\IEEEauthorblockA{\hspace{2cm}\textit{Computer Science and Engineering} \\
\hspace{2cm}\textit{Shahjalal University of Science and Technology}\\
\hspace{2cm}Sylhet, Bangladesh \\
\hspace{2cm}tarifulislamfahim12@gmail.com}
}

\maketitle

\begin{abstract}
Understanding digital documents is like solving a puzzle, especially historical ones. Document Layout Analysis (DLA) helps with this puzzle by dividing documents into sections like paragraphs, images, and tables. This is crucial for machines to read and understand these documents. In the DL Sprint 2.0 competition, we worked on understanding Bangla documents. We used a dataset called BaDLAD with lots of examples. We trained a special model called Mask R-CNN to help with this understanding. We made this model better by step-by-step hyperparameter tuning, and we achieved a good dice score of 0.889. However, not everything went perfectly. We tried using a model trained for English documents, but it didn't fit well with Bangla. This showed us that each language has its own challenges. Our solution for the DL Sprint 2.0 is publicly available at https://www.kaggle.com/competitions/dlsprint2/discussion/432201 along with notebooks, weights, and inference notebook.

\end{abstract}


\section{Introduction}

Deciphering the structure of intricate digital documents is a fundamental stride in transforming them into intelligible machine-readable formats, pivotal for practical applications. With the burgeoning advances in machine learning and the realm of deep neural networks, the task of deciphering and transcribing documents, especially those of historical significance, remains an intricate conundrum. This is where Document Layout Analysis (DLA) emerges as a beacon of understanding. DLA is a preprocessing phase of document transcription. It segments a document into semantic units such as paragraphs, text-boxes, images, and tables. This is essential for OCR, as it allows the OCR engine to correctly identify and extract text from documents.

The DL Sprint 2.0 competition \cite{dlsprint2} was a challenge to develop a DLA system for Bangla documents. The competition dataset, BaDLAD\cite{shihab2023badlad}, contains 33,695 human-annotated document samples from six domains. 

We approached this challenge by training a Mask R-CNN model\cite{8237584} for instance segmentation. Mask R-CNN is a state-of-the-art object detection model that can be used to segment objects in images. We fine-tuned the Mask R-CNN model on the BaDLAD dataset, and we also used hyperparameter optimization to improve the performance of our model.

Our final submission achieved a dice score of 0.88900. This is a competitive result, and it shows that our approach is a promising step toward developing a robust DLA system for Bangla documents.

In addition to our work on the Mask R-CNN model, we also experimented with using pre-trained weights from an English document layout analysis model. However, this did not yield a significant improvement in performance. We believe that this is because the English document layout analysis model was not trained on a dataset that is representative of the challenges of Bangla document layout analysis.

\section{Methodology}

\subsection{Dataset Overview}
In order to improve the performance of Bengali document layout analysis, a multi-domain large Bengali Document Layout Analysis Dataset: BaDLAD has been used to train on the respective model. The dataset contains 3,695 human-annotated document samples from six domains - i) books and magazines ii) public domain govt. documents iii) liberation war documents iv) new newspapers v) historical newspapers and vi) property deeds; with 710K polygon annotations for four unit types: text-box, paragraph, image, and table.

\subsection{Model Overview}
For instance, segmenting objects within images Mask R-CNN R 50 FPN 3x model architecture has been used 
following COCO (Common Objects in Context) format. For efficiency and effectiveness in image analysis tasks,
ResNet-50 backbone architecture is used along with Feature Pyramid Network, assisting the model for better understanding objects at different scales within an image. Moreover, the model is trained for three times the standard number of epochs that are typically used during the training phase. Typically Mask R-CNN (M R-CNN) models are trained for 10000 to 50000 iterations.

\subsection{Tuned Hyperparameters Synopsis}
 \subsubsection{Base Learning Rate}
  The base learning rate, determines the step size of gradient descent when the model updates its parameters during model training. 
  \subsubsection{Pretrained Weights}
  The initial weights for the model to start training from. The model architecture loads the pre-trained model weights if mentioned, instead of untrained initialized weights.
  \subsubsection{Maximum Iteration}
  The maximum number of training iterations. The training process will stop after executing the specified maximum number of iterations.
  \subsubsection{Gamma}
  A factor by which the learning rate is multiplied after each step. It Controls the rate of learning rate decay after specified steps.
  \subsubsection{Warmup Iterations}
  The number of warm-up iterations at the beginning of training where the learning rate gradually increases from a lower value to the base learning rate helps to avoid instability while training.
  
\subsection{Other Hyperparameters}
  \subsubsection{Batch Size}
  In the mini-batch, the gradient descent is applied based on each batch. We have used the mini-batch method with a batch size of 8 images per batch.
  \subsubsection{Workers}
  We have utilized 2 workers to run two batches in parallel.

\subsection{Performance Metric}
\label{sec:performance}

In this section, we discuss the performance metric used to evaluate the effectiveness of our method. The Dice score is a commonly used performance metric for measuring the similarity between two sets. It is often used in medical image segmentation tasks to quantify the agreement between the predicted and ground truth segmentations.

The Dice score (also known as the F1 score or the Sørensen-Dice coefficient) is defined as follows:

\begin{equation}
\label{eq_dice}
\text{Dice Score} = \frac{2 \times \text{Intersection}}{\text{Total Predicted} + \text{Total Ground Truth}}
\end{equation}

Where:
\begin{align*}
\text{Intersection} &= \text{Number of overlapping pixels in} \\
&\quad \text{predicted and ground truth} \\
&\quad \text{segmentations}\\
\text{Total Predicted} &= \text{Total number of pixels} \\
&\quad \text{in predicted segmentation} \\
\text{Total Ground Truth} &= \text{Total number of pixels} \\
&\quad \text{in ground truth segmentation}
\end{align*}

The Dice score ranges from 0 to 1, with 1 indicating perfect overlap between the predicted and ground truth segmentations, and 0 indicating no overlap.
\\
\\
Dice score provides a meaningful measure of segmentation accuracy, allowing us to assess the quality of our method's predictions. Although dice score is a good measure for instance segmentation tasks of computer vision but it still has limitations in certain situations.

 \subsection{Final Sequential Submission Approach}
\begin{table}[h]
\centering
\begin{tabular}{|c|c|c|c|c|c|c|c|}
\hline
\textbf{Sub.} & \textbf{Pretrained Weight} & \textbf{Tr. Split} & \textbf{B. LR} & \textbf{Warmup Iter.} \\
\hline
1 & None & 75\% & 0.007 & 100 \\
2 & Output from sub. 1 & 75\% & 0.001 & 100 \\
3 & Output from sub. 2 & 75\% & 0.0005 & 0 \\
4 & Output from sub. 3 & 75\% & 0.00001 & 0 \\
5 & Output from sub. 4 & 75\% & 0.000001 & 0 \\
6 & Output from sub. 5 & 99\% & 0.000001 & 0 \\
\hline
\end{tabular}
\caption{Training Configuration Part 1: Showing training hyperparameters pre-trained weights, train dataset percentage in train-test split(Tr. Split), Base learning Rate(B. LR), and initial warmup iterations(Warmup Iter.) for each submission Sequence(sub.)}
\label{tab:training-config1}
\end{table}
\begin{table}[h]
\centering
\begin{tabular}{|c|c|c|c|}
\hline
\textbf{Sub.} & \textbf{No. of Iter.} & \textbf{Gamma} & \textbf{Dice Score} \\
\hline
1 & 22k & 0.0001 & 0.88223 \\
2 & 22k & 0.0001 & 0.88783 \\
3 & 22k & 0.0001 & 0.88842 \\
4 & 22k & 0.00001 & 0.8887 \\
5 & 22k & 0.00001 & 0.88894 \\
6 & 5k & 0.00001 & 0.88900 \\
\hline
\end{tabular}
\caption{Training Configuration Part 2: Showing training hyperparameters number of iterations(No. of Iter.) and gamma for each submission Sequence(sub.) and corresponding dice score in percentage(Accuracy)}
\label{tab:training-config2}
\end{table}
Our final submission sequence is decomposed into 6 steps, each of which has been tinkered with different hyperparameter values in order to improve the performance of the respective model on the provided dataset.
\\

In terms of the first submission of the final submission sequence, no pre-trained base model is injected for pre-trained weights. Rather, training commenced from scratch by considering a training split of 75\%, a base learning rate of 0.007, a warmup iteration value of 100, a gamma value of .0001, and a maximum iteration of 22000 eventually providing dice score of 0.88223. \\

Starting from the second submission sequence till the final one, a cumulative approach has been taken in terms of using pre-trained weights. In the case of the second submission, the first submission's pre-trained model weight is injected along with one particular change for the base learning rate having the value of 0.001 and keeping every parameter the same as the previous one providing a dice score of 0.88783, huge improvements over the former one.
\\

The warmup iterations value has been considered from 100 to 0 from the 3rd submission sequence.
For the 3rd, 4th and 5th sequence of submissions, a similar cumulative approach has been applied for adding pre-trained weights of previous submissions to continue its training in the current submissions with updated hyperparameters. Only the learning rate has been different in these submissions eventually providing a dice score of 0.88894. 
\\

Finally, in the last submission, a total of 110,000 iteration version of the model weights has been considered to continue training with a maximum iteration value of 5000 more iterations, train split of 99\% and learning rate of 0.000001. So the final version of the submission is the result of the model being iterated over 115,000 times eventually providing the best dice score for our latest submission, which is 0.88900.

\section{Results and Analysis}
In this section, we will discuss and analyze our results with respect to the hyperparameters tuned at each step of the training. We tested our model on the test data provided by DL sprint 2.0 based on the dice score from the equation (\ref{eq_dice}). The results for each submission are given in the table \ref{tab:training-config2}.
\\

At the first step of our training, we tried training a model from scratch on the DL Sprint 2.0 dataset and training a Mask-RCNN model pre-trained on the PubLayNet dataset \cite{pretrainedgithubrepo}. Although using the pre-trained model gave a better score for the first 10,000 iterations it did not yield any better results on further training even a total of 22, 000 iterations compared to the trained model from scratch. The trained model from scratch for 22k iterations obtains a score of 0.88223 which we selected as our model for further training.
\\

For the second step, as the model showed a high variation of performance at each iteration as we continued to increase our training iterations, so we decreased the initial learning rate to 0.001. Thus training for a further 22,000 iterations yielded a score of 0.88783.
\\

Continuing to our third step, as we continued training the model further on small learning rates the model continued to show slight improvement of score compared to larger similar learning rates of previous steps. So, we tried to keep the learning rate smaller at 0.0005 in this step. As larger learning rates showed variation in scores, we changed the number of warmup steps to 0, so that the learning rate does not increase with iterations initially. In this way, after training for further 22,000 iterations, we got a score of 0.88842.
\\

For the fourth and fifth steps, we continued to observe the same behavior of the learning rate causing us to decrease it further to 0.00001 for the fourth step and 0.000001 for the fifth step. As the learning rate is becoming too low and the model is showing stable improvements on further training, to keep the learning rate stable we decreased the gamma to 10 times making it 0.00001 for both steps. We continued training our model for 22,000 iterations on the fourth step and further for 22,000 iterations on the fifth step resulting in a score of 0.8887 and 0.88894 respectively.
\\

In the sixth step of our continuation of training, we noticed a stable improvement in model performance just by increasing the training data making the training data to validation data ratio of 99:1 compared to the previous 75\% of our training data. With this hyperparameter change and continuing our training for 5,000 iterations lead our model to improve to a score of 0.889.
\\

In total, we trained our model for 115k iterations with varying hyperparameters at each step reaching a model performance dice score of 0.889. We can see that the model training shows a gradual increase in performance with increasing iterations and selecting proper hyperparameters after specific iterations as depicted in the table \ref{tab:training-config2}. We can also see that increasing training data after some iteration increases the model performance.

\section{Conclusion}

In this paper, we presented our approach to the DL Sprint 2.0 competition. We trained a Mask R-CNN model for instance segmentation on the BaDLAD dataset, and we also used hyperparameter optimization to improve the performance of our model. Our final submission achieved a dice score of 0.889.

We believe that our approach is a promising step towards developing a robust DLA system for Bangla documents. However, there is still more work to be done. In the future, we plan to improve our approach by enhancing the dataset and by incorporating more advanced techniques. 

We hope that our work has the potential to add value to the field of document layout analysis. DLA is a key technology for many applications, such as OCR, machine translation, and search. By developing better DLA systems for Bangla documents, we can make these applications more accessible to the millions of people who speak Bangla.

We have observed that using proper hyperparameters can improve the model's performance with further training. Also increasing the dataset also increases the performance of the model.

\vspace{12pt}

    \bibliographystyle{plain}
    \bibliography{reference}
    \nocite{dlsprint2}
    \nocite{shihab2023badlad}
    \nocite{pretrainedgithubrepo}
    \nocite{8237584}


\end{document}